\documentclass[conference]{IEEEtran}
\IEEEoverridecommandlockouts
\usepackage{cite}
\usepackage{amsmath,amssymb,amsfonts}
\usepackage{algorithmic}
\usepackage{graphicx}
\usepackage{textcomp}
\usepackage{xcolor}
\usepackage{times}
\usepackage{xcolor}
\usepackage{soul}
\usepackage[utf8]{inputenc}
\usepackage[small]{caption}
\usepackage{amsmath}
\usepackage{graphicx}
\usepackage[colorinlistoftodos]{todonotes}
\usepackage{algorithm}
\usepackage{algorithmic}
\usepackage{amssymb}
\usepackage{float}
\usepackage{graphicx}
\usepackage{subcaption}
\usepackage{tikz}
\usepackage{pgf-pie}
\usepackage{multirow}
\usepackage{hhline}
\def\BibTeX{{\rm B\kern-.05em{\sc i\kern-.025em b}\kern-.08em T\kern-.1667em\lower.7ex\hbox{E}\kern-.125emX}}

\begin{document}

\title{Ensemble-based Adaptive Single-shot Multi-box Detector}


\author{
\IEEEauthorblockN{Viral Thakar\IEEEauthorrefmark{1}, Walid Ahmed\IEEEauthorrefmark{2}, Mohammad M Soltani\IEEEauthorrefmark{2}, Jia Yuan Yu\IEEEauthorrefmark{3}}

\IEEEauthorblockA{\IEEEauthorrefmark{1}Department of Electrical and Computer Engineering\\Concordia University, Montreal, Canada\\Email: v\_thakar@encs.concordia.ca}
\IEEEauthorblockA{\IEEEauthorrefmark{2}Indus.ai, Thornhill, Canada\\
Email: walid.aly@indus.ai, mohammad.soltani@indus.ai}
\IEEEauthorblockA{\IEEEauthorrefmark{3}Concordia Institute of Information System Engineering\\Concordia University, Montreal, Canada\\Email: jiayuan.yu@concordia.ca}}

\maketitle

\begin{abstract}
We propose two improvements to the SSD---single shot multibox detector. First, we propose an adaptive approach for default box selection in SSD. This uses data to reduce the uncertainty in the selection of best aspect ratios for the default boxes and improves performance of SSD for datasets containing small and complex objects (e.g., equipments at construction sites). We do so by finding the distribution of aspect ratios of the given training dataset, and then choosing representative values. Secondly, we propose an ensemble algorithm, using SSD as components, which improves the performance of SSD, especially for small amount of training datasets. 
Compared to the conventional SSD algorithm, adaptive box selection improves mean average precision by 3\%, while ensemble-based SSD improves it by 8\%.
\end{abstract}


\section{Introduction}
Object Detection is a domain which has fascinated computer vision researchers since the beginning. It involves the localization and classification of objects available in an image or video. Deep object detectors can be divided into two categories: (i) Two-stage approaches which applies a classifier on a sparse set of proposed candidate object locations (ii) Single-stage approaches which divide the input space into dense set of boxes called default boxes and apply a classifier on them. Performance wise, it is always been found that two stage approaches \cite{DBLP:journals/corr/HeGDG17,DBLP:journals/corr/RenHG015} are better in terms of accuracy compared to single-stage networks because of their localization capabilities. However in hindsight, two stage approaches are comparatively slower. On the other hand, in recent work \cite{DBLP:journals/corr/LiuAESR15,DBLP:journals/corr/RedmonF16,DBLP:journals/corr/FuLRTB17} the single-stage approaches have also shown a potential performance in terms of speed and accuracy \cite{DBLP:journals/corr/HuangRSZKFFWSG016}. The purpose of this research is to push the capabilities of single-stage networks by improving their detection precision.  

SSD - Single Shot Multibox Detector \cite{DBLP:journals/corr/LiuAESR15} belongs to the family of object detection algorithms which uses single deep neural network to detect different object classes. For detecting objects, rather hypothesizing bounding boxes or re-sampling pixels or features for each box and then applying a high quality classifier; SSD discretized the output space of bounding boxes into a set of default boxes over different aspect ratios and scales per feature map location. It then generates scores for the presence of each object class in each default box and produces adjustments to better match object shape. The fundamental concept of SSD is mostly based on the feed forward convolution network. The SSD model is comprised of mainly two structures : Base network and Auxiliary network. The Base network is the early part of the model which is based on standard architecture used for high quality image classification. The Auxiliary network has features mainly focused for objects with different scales or aspect ratios. 

There have been various approaches \cite{DBLP:journals/corr/JeongPK17,DBLP:journals/corr/abs-1709-05054,DBLP:journals/corr/FuLRTB17} proposed to improve the performance of SSD by enhancing its capabilities to collect more features. In \cite{DBLP:journals/corr/HuangRSZKFFWSG016}, the authors have shown that the performance of SSD is always very low for small objects no mater how much deep feature extractor we use. It was also found that the performance of SSD degrades more when the training dataset contains small, complex and deforming objects. Its robustness for real life applications is a point of concern. Motivated by these limitations and SSD's popularity because of its capability to operate in real-time, we wish to find the possible improvements to SSD which can allow us to augment its performance for small and complex object detection. 

SSD generates a set of default boxes based on a set of static values for aspect ratio. For any type of training data, the generation of default boxes is fixed and never changes. The problem with the fixed set of aspect ratios and scales is that the generated default boxes are not ideal for the given training dataset. This may end up in having either multiple or very poor localization over the same object, in turn adding to false positives. By having an adaptive default box generation algorithm which creates a set of default boxes based on the distribution of aspect ratios across the training data, we can optimize the selection of the default boxes. Consequently, this creates better localization and classification.

\IEEEpubidadjcol

The paper is organized as follows. Section 2 describes the model architecture and training process of SSD. Next, we present in two section our main contributions to improve the performance of SSD. Section 3 covers the adaptive default box selection algorithm for SSD. Section 4 explains the ensemble of SSDs and process to train the ensemble of SSDs. Performance comparison of different improvements to SSD with conventional SSD is discussed in Section 5. Finally we present conclusion and open questions in Section 6.

\section{SSD - Single Shot Multi-box Detector}
SSD uses a single-stage, feed-forward convolutional neural network  for object detection. The key contribution of SSD is the use of default boxes over multi-scale feature maps for detection, which are similar to the anchor boxes of Faster RCNN \cite{DBLP:journals/corr/RenHG015}. This allows SSD to discretize the space of possible output box shapes. It takes a set of images containing objects along with object labels and boxes circumscribing this objects. This boxes are called ground-truth boxes. 
To begin the training, it first evaluates a set of default boxes with different aspect ratios and scales at each location in several feature maps. It then predicts both the shape offsets and the confidences for each default box and all object classes. The goal during the training is to match the default boxes with the ground truth boxes for a high class confidence score. To achieve this goal, it uses weighted sum of localization loss and confidence loss as the loss function. 

Mathematically SSD is a function $\varphi(x) = \hat{Y}$ which takes any arbitrary image $x$ as input. For number of classes $n$ and number of default boxes $d$, SSD produces a matrix $\hat{Y} \in \mathbb{R}^{d \times (n + 4)}$ as output. Each row of $\hat{Y}$ represents a positive real valued vector $\hat{y} \in \mathbb{R}^{n+4}$. It contains $n$ real valued numbers representing per-class classification probabilities or confidences and four real valued numbers to represent the offset in the default box. For simplicity consider a matrix $\hat{Y}_{clss} \in \mathbb{R}^{d \times n}$ which represents only classification probabilities for $d$ default boxes and a matrix $\hat{Y}_{loc} \in \mathbb{R}^{d \times 4}$ which represents the offset in the $d$ default boxes. 

The training of SSD on single training sample can be divided into three main sections. Repeating the same key steps for each training sample can lead to the full training approach used by SSD.
\begin{algorithm}
   \caption{Initialize Set of Default Boxes D}
   \label{alg1}
\begin{algorithmic}
	\STATE{\bfseries Inputs:}
   		\\$m$ - Number of Feature maps locations for prediction
        \\$s_{min}$ - Minimum Scale Value - Default 0.2
        \\$s_{max}$ - Maximum Scale Value - Default 0.9
   	\STATE{\bfseries Output:}
    	\\$D \in \mathbb{R}^{d\times4}$ - Set of Default Boxes
    \STATE{\bfseries Process:}
    		\\Initialize Scale of Default boxes $s \in \mathbb{R}^m$
        	\FOR{each feature map $k \in [1, m]$}
            	\STATE{$s[k] = s_{min} + \frac{s_{max} - s_{min}}{m - 1}(k - 1)$}
            \ENDFOR
         
            \FOR{each aspect ration $a_r \in \{1, 2, 3, 1/2, 1/3\}$}
            	\FOR{each feature map $k \in [1, m]$}
                		\STATE{Size of $k^{th}$ feature map $f_k$}
                        \STATE{$(c_x, c_y) = (\frac{i + 0.5}{f_k}, \frac{j+0.5}{f_k})$} where $i, j \in [0, f_k)$
                        \IF{$a_r == 1$}
                        	\STATE{Width $w{_k^{a_r}} = s_k * \sqrt[]{a_r}$}
                        	\STATE{Height $h{_k^{a_r}} = s_k / \sqrt[]{a_r}$}
                            \STATE{Default Box $D{_k^{a_r}} = [{c_x, c_y, w{_k^{a_r}}, h{_k^{a_r}}}]$}
                        	\STATE{$s{^{'} _k} = \sqrt[]{s_k * s_{k+1}}$}
                            \STATE{Width $w{_k^{'{a_r}}} = s{^{'} _k} * \sqrt[]{a_r}$}
                        	\STATE{Height $h{_k^{'{a_r}}} = s{^{'} _k} / \sqrt[]{a_r}$}
                            \STATE{Default Box $D{_k^{{'a_r}}} = [{c_x, c_y, w{_k^{'{a_r}}}, h{_k^{'{a_r}}}}]$}
                        
                        \ELSE
                        	\STATE{Width $w{_k^{a_r}} = s_k * \sqrt[]{a_r}$}
                        	\STATE{Height $h{_k^{a_r}} = s_k / \sqrt[]{a_r}$}
                            \STATE{Default Box $D{_k^{a_r}} = [{c_x, c_y, w{_k^{a_r}}, h{_k^{a_r}}}]$}
                        \ENDIF
                \ENDFOR 
            \ENDFOR
\end{algorithmic}
\end{algorithm}

\subsection{Initializing Default Boxes}
The default box selection is based on the minimum and maximum scale values, as well as the number of feature maps needed to be used for the prediction. The algorithm for default box selection is motivated from \cite{DBLP:journals/corr/LongSD14,DBLP:journals/corr/HariharanAGM14a,DBLP:journals/corr/LiuRB15} which are object segmentation algorithms. The Algorithm \ref{alg1} shows the process of generating default boxes for a given number of feature maps $m$ and bounding values $s_{min}$ and $s_{max}$ for scale of boxes. For each feature map $k \in [1, m]$, SSD calculates the scale of the default boxes denoted as $s_k$. This is essentially the equal distribution of scale values between $s_{min}$ and $s_{max}$ based on the number of feature maps. For example, for number of feature maps m = 16, SSD divides the $[s_{min}, s_{max}]$ interval into 16 equally distributed parts. Next for each scale value $s_k$ associated with a particular feature map, SSD considers five aspect ratio values from \{1, 2, 3, 1/2, 1/3\} and creates rectangle boxes with each one. The boxes are centered at $(c_x, c_y)$ point. For the aspect ratio value 1, SSD considers an extra box with scale $s{^{'} _k} = \sqrt[]{s_k * s_{k+1}}$ which makes a total of six default boxes for a particular feature map. 
\begin{algorithm}
   \caption{Create set of Matched Boxes M for Single Training Image }
   \label{alg2}
\begin{algorithmic}
	\STATE{\bfseries Inputs:}
   		\\$\tau$ - Jaccard overlap threshold - Default 0.5
        \\$d$ - Number of default boxes
        \\$g$ - Number of ground truth boxes
        \\$D \in \mathbb{R}^{d\times4}$ - Set of default boxes
        \\$G \in \mathbb{R}^{g\times4}$  - Set of ground truth boxes
        \\$C$ - A set of class labels 
        \\$n$ - Total number of class labels
   	\STATE{\bfseries Output:}
    	\\$N$ - Number of positively matched default boxes
    	\\$Pos \in \mathbb{R}^{N}$  - Indexes of positively matched default boxes
        \\$Neg \in \mathbb{R}^{(d - N)}$ - Indexes of negatively matched default boxes
    \STATE{\bfseries Initialize :}
    	\\$Y_{clss} \in \mathbb{R}^{d \times n}$ - Ground-truth labels for each default box
        \\$Y_{loc} \in \mathbb{R}^{N \times 4}$ - Ground-truth boxes for each positively matched default box
    \STATE{\bfseries Process:}
        	\FOR{each $i^{th}$ default box $D[i]$}
            	\FOR{each $j^{th}$ ground truth box $G[j]$ having $c^{th}$ class label $C[c]$}
                    \STATE{Jaccard Overlap $J(D[i],G[j]) = 1 - \frac{|G[j] \cap D[i]|}{|G[j] \cup D[i]|}$}
                    \IF{$J(D[i],G[j]) > \tau$}
                        \STATE{$Y_{clss}(i,c) = 1$}
                        \STATE{Append G[j] to $Y_{loc}$}
                        \STATE{Append $i$ to Pos}
                    \ELSE
                        \STATE{$Y_{clss}(i,c) = 0$}
                        \STATE{Append $i$ to Neg}
                    \ENDIF
                  \ENDFOR
            \ENDFOR
\end{algorithmic}
\end{algorithm}

\subsection{Defining a Matching Strategy between Default Boxes and Ground truth Boxes}
The next stage in the training part is to match the default boxes with the ground truth boxes. For a single image, the matching strategy is explained in the Algorithm \ref{alg2}. The idea is to find all the default boxes which are overlapping on the ground truth boxes in order to consider them as a set of positive samples for the classifier training. The default boxes which are not sufficiently overlapping on the ground truth boxes are considered as negative samples or background. After getting a set of positive and negatives boxes, SSD uses hard negative mining \cite{DBLP:journals/corr/ShrivastavaGG16} to balance the number of positive and negative samples. This keeps the ratio of number of negative samples to number of positive samples at 3 : 1. 

\subsection{Defining Training Objective and Loss Function}
The next step is to define the training objective in terms of minimizing loss function. As the object detection task involves classification and localization, the loss function is also a weighted sum of classification and localization loss functions. As mentioned earlier, for each arbitrary image SSD produces a matrix $\hat{Y} \in \mathbb{R}^{d \times (p+4)}$, which represents a set of real valued vectors for each default boxes. For simplicity we have also defined $\hat{Y}_{clss} \in \mathbb{R}^{d \times p}$ and $\hat{Y}_{loc} \in \mathbb{R}^{d \times 4}$ representing the predictions for classification and localization task respectively. 

The classification loss is simply Softmax loss or more accurately known as cross-entropy loss function over multiple classes confidences.

$$L_{clss}(\hat{Y}_{clss}, Y_{clss}) = - \frac{1}{N} \sum_{c=1}^{n} \sum_{l \in Pos} Y_{clss}[l,c] $$
$$\cdot \log(\hat{Y}_{clss}[l,c]) - \sum_{h \in Neg} \log(\hat{Y}_{clss}[h,0]).$$

The localization loss is smooth L1 loss \cite{DBLP:journals/corr/Girshick15} defined as 

$$L_{loc}(\hat{Y}_{loc}, Y_{loc}) = \sum_{b \in (c_x, c_y, w, h)} smooth_{L_1}(\hat{Y}_{loc}[b,:] - Y_{loc}[b,:])$$

where 

$$smooth_{L_1}(x) = \Big\{\begin{tabular}{c} $0.5 \times x^2$ if $|x| > 1$ \\ $|x| - 0.5$ otherwise \end{tabular}.$$

The overall loss function is 
$$L(\hat{Y}, Y) = \frac{1}{N} (L_{clss}(\hat{Y}_{clss},Y_{clss}) + \alpha (L_{loc}(\hat{Y}_{loc},Y_{loc})).$$

\section{Adaptive Default Box Selection Algorithm for SSD}
The performance of two-stage detectors is mainly because of a dedicated region proposal network. Default boxes are the ones which do a similar job like region proposal network of two stage object detectors. This is the motivation to improve the default box selection part of SSD. The selection of default boxes in SSD is dynamic with respect to the scale, but it is static with respect to the aspect ratio. It chooses aspect ratios of default boxes from a fix set of values. Fig. \ref{fig:ardistr} shows the distribution of aspect ratios for the selected dataset mentioned in section 5.1. This clearly shows that each object label in the dataset has different distribution of aspect ratios, thus selecting aspect ratios from ${1, 2, 3, 1/2, 1/3}$ is not an optimized approach. 

Considering the appearance of object shapes throughout the training data as a random process, a positive real valued random variable $X : \Omega \rightarrow \mathbb{R}_{> 0}$ is defined on the probability space $(\Omega, \digamma, P)$. $X$ can map the set of all possible aspect ratio values $\Omega$ to some positive real numbers. For the given training set, we can find the probability density of aspect ratios for a particular object class which can be represented as follows.

$$ F_X(x) = P(X \leq x) = \int_{0}^{x} f_X(u) du $$ 

where $f_X$ is the probability density function of $X$, and $F_X$ is the cumulative distribution function of $X$. The goal is to find five representative points; $\{x_1, x_2, x_3, x_4, x_5\}$ for this density function 

The first representative point $x_1$ can be the mode of $f_X$, which represents a value of $x$ for which $f_X(x)$ takes a maximum value. In other words, mode is the x-coordinate of the maximum point on the graph of $f_X(x)$. The second representative point, $x_2$ can be the mean of $f_X$ given as $$ x_2 = \int_{0}^{inf} xf_X(x) dx.$$ $x_2$  represents the arithmetic mean of the aspect ratios for objects available in the training data. 

The third representative point $x_3$ can be the median of  $f_X$ given as $x_3 = m$ such that $$\int_{0}^{m} f_X(x) dx = \int_{m}^{inf} f_X(x) dx = 0.5.$$ This represents the line $x_3 = m$ dividing the area under the graph of $f_X(x)$ into two equal areas. Next, two representative points are given as $x_4 = m$ such that $$\int_{0}^{m} f_X(x) dx = 0.25$$ and $x_5 = m$ such that $$\int_{0}^{m} f_X(x) dx = 0.75.$$

In our experiment we are estimating the probability density function $f_X$ using histogram as estimation technique. For a given starting point $x_0$ and bin width $h$ and total samples $n$, the bins of the histogram can be established as

$$ [x_0 + \beta h, x_0 + (\beta + 1)h].$$

The histogram estimation is defined as 

$$f'{_X}(x) = \frac{1}{nh} \cdot (\textrm{No of } X_i \textrm{ in same bin as }x).$$

\section{Ensemble of SSDs}
Ensemble methods are methods which generate a set of learning algorithms and then use a voting mechanism among them to predict a output for a new data point. An ensemble of classifiers is a set of classifiers whose individual decisions are combined in some way typically by weighted or unweighted voting to classify new examples. A necessary and sufficient condition for an ensemble of classifiers to be more accurate than any of its individual members, is if the classifiers are accurate and diverse. An accurate classifier is the one which has high accuracy than random guessing, wheras two classifiers are diverse if they make different errors on new data points. More precisely, if the error rates of $L$ hypotheses $\{h_1, h_2, ... h_l\}$ have probability $p < 0.5$ and the errors are independent, then the probability that the majority vote will be wrong is equivalent to the area under the binomial distribution where more than $L/2$ hypotheses are wrong \cite{dietterich2002ensemble}.

\subsection{Bootstrap Aggregating - Bagging}
As SSD uses some similar approach like boosting \cite{freund1996experiments} by implementing hard negative mining, we have decided to explore the effect of bagging \cite{Breiman1996,breiman1996heuristics} as the combination of SSDs. It is a method for generating multiple versions of a specific predictor and using these to get an aggregated predictor. To generate aggregation, either take the average over the versions while attempting to solve a regression problem, or take a plurality vote among the versions while solving a classification problem. The multiple versions are formed by generating bootstrap replicates of original learning set and use them as individual new learning sets. The vital element for bagging is the instability of the prediction method. If altering the data changes the parameters of the constructed predictor, bagging can improve accuracy. 

\subsubsection{Problem Statement for Bagging}
Consider a training data set $L$, consists of data $\{ (\textbf{x}_n, y_n), n = 1, 2, ... N \}$ where $y$ is the label and $\textbf{x}$ is a feature vector. We have some learning system to form a predictor $\varphi(\textbf{x}, L)$ which predicts $y$ for any new sample of $\textbf{x}$. Now let's consider we are having a sequence of learning sets, ${L_k}$ each consisting of $N$ independent observations from the same underlying distribution as $L$. Our primary goal is to use $\{L_k\}$ to built a better predictor than the single learning set predictor $\varphi(\textbf{x}, L)$. The only restriction is that we have to work with a sequence of predictors $\{\varphi(\textbf{x}, L_k)\}$ \cite{Breiman1996}.

The obvious approach for the classification problem, i.e. to predict class labels, is voting. Consider a set of classes $Y = \{ 1, 2, ... J\}$ and our predictor predicts a class $j \in Y$ then aggregated predictor is given as: $$\varphi_A(\textbf{x}) = argmax_j V_j$$ where $V_j = \#\{k; \{\varphi(\textbf{x}, L_k)\} = j \}$. 

Usually we have a single training set $L$ and not a sequence of training sets $\{L_k\}$, we can take repeated bootstrap samples and create a sequence $\{L^{(B)}\}$ from $L$ and form sequence of predictors $\{\varphi(\textbf{x}, L^{(B)})\}$. $\{L^{(B)}\}$ creates replicate datasets, each having $N$ training samples drawn at random, but with replacement from $L$. Each pair $(\textbf{x}, y)$ from training set $ L = \{ (\textbf{x}_n, y_n), n = 1, 2, ... N \}$ may appear repeated times or not at all in any particular $L^{(B)}$. 

The answer to one of the most critical questions about bagging that whether it will improve accuracy or not depends upon the stability of the procedure for the construction of our predictor $\varphi$. If the small change in $L$ creates a large change in the parameters of $\varphi$ then bagging insures the improvements in performance. As suggested in \cite{breiman1996heuristics} neural networks, classification and regression trees are unstable methods while k-nearest neighbor method is stable. This provides a motivation to explore the bagging for Object Detection based on the Convolutional Neural Networks. 

\subsubsection{Procedure}
\begin{enumerate}
\item The dataset is randomly divided into a test set $T$ and training set $L$. Usually we divide 20\% of dataset for testing set $T$ and 80\% of dataset for training set $L$. 

\item An Object Detector e.g. SSD $\varphi(\textbf{x})$ is trained using $L$. Evaluating the test set $T$ through Object Detection algorithm gives the performance parameter Mean Average Precision denoted as $mAP_s$. 

\item A bootstrap sample $L^{(B)}$ is selected from $L$ and Object Detector SSD is trained using $L^{(B)}$. This step is repeated $m$ times giving Object Detectors $\varphi_1(\textbf{x}), ... , \varphi_m(\textbf{x})$

\item If $(\textbf{x}_n, y_n) \in T$, then the estimated label of $\textbf{x}_n$ is that label having the plurality in $\varphi_1(\textbf{x}_n), ... , \varphi_m(\textbf{x}_n)$. The performance parameter Mean Average Precision denoted as $mAP_B$ can be calculated. 
\end{enumerate}

\section{Experiment Setup and Results}

\begin{figure}
    \centering
    \begin{subfigure}[b]{0.45\linewidth}
        \includegraphics[width=\textwidth, height=1.5cm]{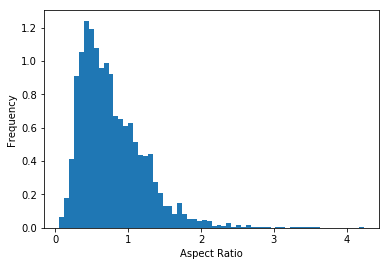}
        \caption{Equipment-1}
        \label{fig:bus}
    \end{subfigure}
    ~
    \begin{subfigure}[b]{0.45\linewidth}
        \includegraphics[width=\textwidth, height=1.5cm]{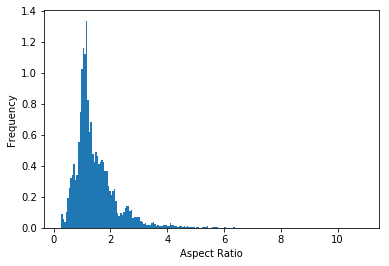}
        \caption{Equipment-2}
        \label{fig:drillrig}
    \end{subfigure}
	\\
    \begin{subfigure}[b]{0.45\linewidth}
        \includegraphics[width=\textwidth, height=1.5cm]{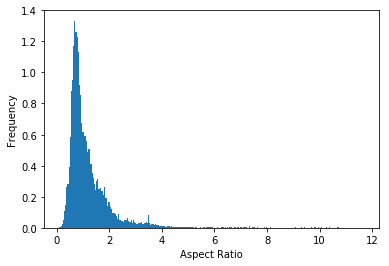}
        \caption{Equipment-3}
        \label{fig:equipment}
    \end{subfigure}
    ~
    \begin{subfigure}[b]{0.45\linewidth}
        \includegraphics[width=\textwidth, height=1.5cm]{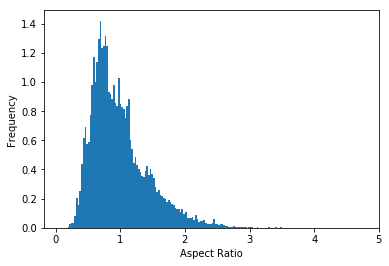}
        \caption{Equipment-4}
        \label{fig:excavator}
    \end{subfigure}
	\\
    \begin{subfigure}[b]{0.45\linewidth}
        \includegraphics[width=\textwidth, height=1.5cm]{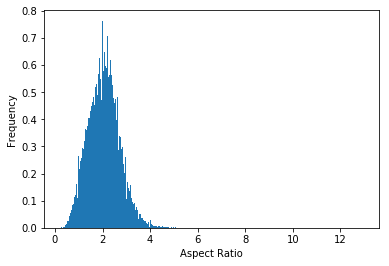}
        \caption{Equipment-5}
        \label{fig:person}
    \end{subfigure}
    ~
    \begin{subfigure}[b]{0.45\linewidth}
        \includegraphics[width=\textwidth, height=1.5cm]{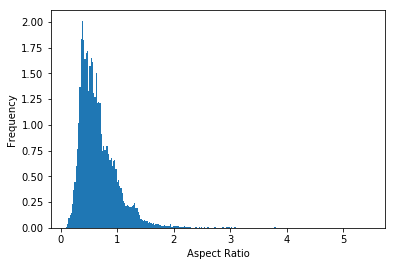}
        \caption{Equipment-6}
        \label{fig:truck}
    \end{subfigure}
	\\
    \begin{subfigure}[b]{0.45\linewidth}
        \includegraphics[width=\textwidth, height=1.5cm]{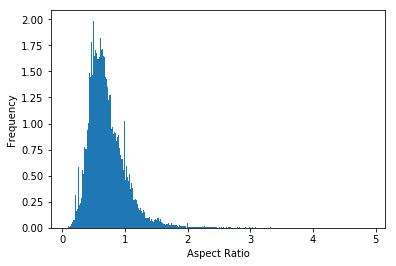}
        \caption{Equipment-7}
        \label{fig:vehicle}
    \end{subfigure}
    ~
    \begin{subfigure}[b]{0.45\linewidth}
        \includegraphics[width=\textwidth, height=1.5cm]{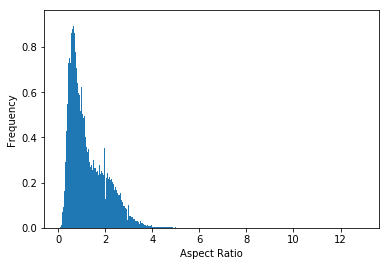}
        \caption{All Data}
        \label{fig:alldata}
    \end{subfigure}
    \caption{Aspect Ratio Distribution of all the classes}\label{fig:ardistr}
\end{figure}

\begin{figure}
\begin{tikzpicture}
\pie [radius=2.4, color={black!5, black!10, black!15, black!20, black!25, black!30, black!35}]{1.99/Equipment-1, 3.30/Equipment-2, 8.62/Equipment-3, 6.33/Equipment-4, 22.87/Equipment-5, 26.14/Equipment-6, 30.75/Equipment-7} 
\end{tikzpicture}
\caption{Testing Dataset Statistics}\label{fig:testdatastats}
\end{figure}
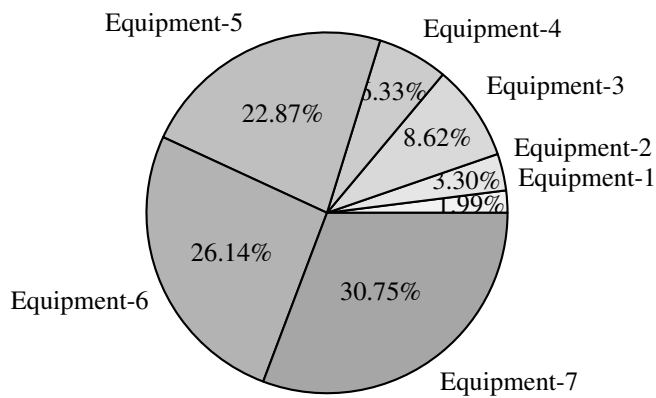

As the goal of this research is to improve the performance of SSD for small objects present in complex environments like construction sites, we have created the dataset from images taken from real construction sites having a different range of objects from excavators to workers. The dataset has seven labels : Equipment-1 to Equipment-7. The images are taken from surveillance cameras placed at various construction sites with different angles and heights. This setup allows us to generate a real dataset with objects ranging in different scales and different aspect ratios. Fig. \ref{fig:ardistr} shows the distribution of aspect ratios of objects present in the dataset. Fig. \ref{fig:traindatastats} and \ref{fig:testdatastats} shows the details about training and testing dataset. 

To understand and evaluate the performance of our proposed improvements to SSD, we have considered two variants of SSD. i) SSD with Inception \cite{DBLP:journals/corr/SzegedyVISW15} as base network ii) SSD with Mobilenet \cite{DBLP:journals/corr/HowardZCKWWAA17} as base network. To evaluate the performance we have chosen mean average precision \cite{everingham2010pascal} as standard performance matrix for object detection. During this research we have evaluated following results.

As a first step, we have trained SSD with the training data mentioned in section 5.1. While evaluating performance of SSD we have found that it performs decently for objects which appear large in terms of scale and aspect ratio. To be fair in evaluation, while creating different bags we have created bootstrap samples by randomly selecting 80\% samples from the available training data. So each version of SSD trained with Bag\_0, Bag\_1 and Bag\_2 is actually trained with less data. In general we refer these SSDs as Bagged SSDs. We have also trained one version of SSD with the adaptive default box selection algorithm and called it ADBS-SSD.

The next step is the evaluation of different approaches. Tables \ref{inceptiontable} and \ref{mobilenettable} show the performance comparison of different experiments carried out as a part of this research. They show the comparison of Average Precision for each class and mean Average Precision as overall performance for different approaches. It clearly shows that the performance of SSD for classes like Equipment-2 and Equipment-4 is better but it struggles for Equipment-5 detection as appearance of Equipment-5 in surveillance camera is comparatively small. First we have evaluated ADBS-SSD with Inception as base network on the test dataset and found maximum 9\% of improvement in average precision for class Equipment-3 and minimum 3\% improvement in average precision for class Equipment-1. Similarly the evaluation of ADBS-SSD with MobileNet as base network has also shown improvements in individual average precision as well as mAP. The performance of Equipment-5 detection is also improved by 4\% by using ADBS-SSD with Inception as well as MobileNet as base network. 

\begin{figure}
\begin{tikzpicture}
\pie [radius=2.4, color={black!5, black!10, black!15, black!20, black!25, black!30, black!35}]{1.81/Equipment-1, 5.87/Equipment-2, 13.44/Equipment-3, 11.54/Equipment-4, 28.74/Equipment-5, 16.28/Equipment-6, 22.32/Equipment-7} 
\end{tikzpicture}
\caption{Training Dataset Statistics}\label{fig:traindatastats}
\end{figure}
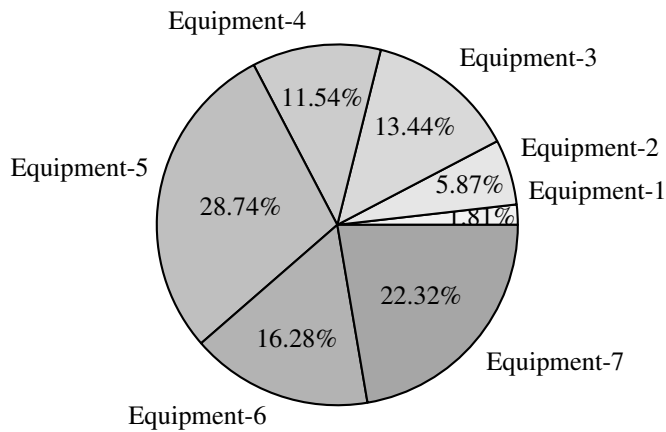

\begin{table*}
\centering
\caption{Average Precision per Class and Mean Average Precision (mAP) Comparison for Inception as Base Network}
\label{inceptiontable}
\begin{tabular}{|c|c|c|c|c|c|c|c|c|}
\hline
\multirow{2}{*}{\textbf{Class}} & \multirow{2}{*}{\textbf{Bag 0}} & \multirow{2}{*}{\textbf{Bag 1}} & \multirow{2}{*}{\textbf{Bag 2}} & \multirow{2}{*}{\textbf{SSD}} & \multicolumn{3}{c|}{\textbf{Bagged SSD}}	& \multirow{2}{*}{\textbf{ADBS SSD}} \\ \cline{6-8}	& 		&		&		&		& \textbf{SSD + Top Bag} & \textbf{Top 2 Bags} & \textbf{All Bags} &	\\ \hline
Equipment-1		& 0.22		& 0.23		& 0.21		& 0.24		& 0.28		& 0.33		& 0.31		& 0.24\\ \hline
Equipment-2		& 0.86		& 0.84		& 0.88		& 0.87		& 0.87		& 0.90		& 0.91		& 0.88\\ \hline
Equipment-3		& 0.55		& 0.55		& 0.56		& 0.50		& 0.55		& 0.66		& 0.61		& 0.61\\ \hline
Equipment-4		& 0.92		& 0.91		& 0.90		& 0.90		& 0.91		& 0.94		& 0.94		& 0.93\\ \hline
Equipment-5		& 0.17		& 0.18		& 0.20		& 0.20		& 0.23		& 0.29		& 0.30		& 0.23\\ \hline
Equipment-6		& 0.45		& 0.41		& 0.46		& 0.48		& 0.50		& 0.53		& 0.58		& 0.54\\ \hline
Equipment-7		& 0.32		& 0.34		& 0.36		& 0.34		& 0.38		& 0.48		& 0.45		& 0.37\\ \hline
mAP				& 0.50		& 0.49		& 0.51		& 0.51		& 0.53		& 0.59		& 0.59		& 0.54\\ \hline
\% Improvement	& \multicolumn{4}{c|}{}							& 2.0		& 8.0		& 8.0		& 3.0\\ \hline
\end{tabular}
\end{table*}

\begin{table*}
\centering
\caption{Average Precision per Class and Mean Average Precision (mAP) Comparison for MobileNet as Base Network}
\label{mobilenettable}
\begin{tabular}{|c|c|c|c|c|c|c|c|c|}
\hline
\multirow{2}{*}{\textbf{Class}} & \multirow{2}{*}{\textbf{Bag 0}} & \multirow{2}{*}{\textbf{Bag 1}} & \multirow{2}{*}{\textbf{Bag 2}} & \multirow{2}{*}{\textbf{SSD}} & \multicolumn{3}{c|}{\textbf{Bagged SSD}}       & \multirow{2}{*}{\textbf{ADBS SSD}} \\ \cline{6-8}
                       &                        &                        &                        &                      & \textbf{SSD + Top Bag} & \textbf{Top 2 Bags} & \textbf{All Bags} &                           \\ \hline
Equipment-1		& 0.16		& 0.28		& 0.20		& 0.23		& 0.24		& 0.27		& 0.25		& 0.22\\ \hline
Equipment-2		& 0.77		& 0.84		& 0.66		& 0.61		& 0.63		& 0.67		& 0.64		& 0.65\\ \hline
Equipment-3		& 0.32		& 0.46		& 0.37		& 0.40		& 0.38		& 0.44		& 0.47		& 0.40\\ \hline
Equipment-4		& 0.92		& 0.89		& 0.82		& 0.87		& 0.88		& 0.91		& 0.88		& 0.87\\ \hline
Equipment-5		& 0.12		& 0.16		& 0.14		& 0.12		& 0.14		& 0.18		& 0.19		& 0.16\\ \hline
Equipment-6		& 0.29		& 0.37		& 0.25		& 0.26		& 0.25		& 0.36		& 0.30		& 0.28\\ \hline
Equipment-7		& 0.22		& 0.31		& 0.26		& 0.23		& 0.28		& 0.31		& 0.25		& 0.24\\ \hline
mAP				& 0.40		& 0.47		& 0.39		& 0.39		& 0.40		& 0.45		& 0.43		& 0.40\\ \hline
\% Improvement	& \multicolumn{4}{c|}{}							& 1.0		& 6.0		& 4.0		& 1.0\\ \hline
\end{tabular}
\end{table*}

To evaluate the bagging, we carried out three experiments. i) We have considered the voting between SSD and Bagged SSD with maximum mAP i.e. SSD trained on Bag 2 for Inception base network and SSD trained on Bag 1 for MobileNet base network. ii) We have considered the voting between top-2 Bagged SSDs. iii) We have considered voting between all the bags. The purpose of carrying out these three experiments is to understand what should be the optimal number of bags, as each bagged SSD in evaluation reduces the FPS of overall performance. By this we can understand more about speed-accuracy trade off associated with bagging. The FPS of SSD is 40 while testing on NVIDIA GTX GeForce 1060 which reduces to almost 10 FPS while using all the bags. For combination of top two bags of SSD we are getting 24 FPS which is sufficient enough for real-time object detection. 

\section{Conclusion}
In this paper, we propose two improvements to SSD in terms of Adaptive Default Box Selection algorithm and Ensemble of SSDs. By using optimal values of aspect ratios for default box generation we are able to improve the performance of SSD for small objects like Equipment-5 detection in surveillance cameras. Use of ensemble of SSDs make the detection more accurate for custom objects in complex surrounding like construction site resources. Quantitative evaluation of performance of each improvements shows that they improves the accuracy while maintaining the FPS required for real-time application. Both the improvements, adaptive default box selection algorithm and ensemble of object detectors, are easy to implement can be extended to any single stage object detector.  

\section*{Acknowledgment}

We acknowledge the support of the Natural Sciences and Engineering Research Council of Canada (NSERC), [funding reference number 396151363].

\bibliographystyle{IEEEtran}
\bibliography{IEEEabrv,references.bib}

\end{document}